\documentclass[sigconf,screen,authorversion]{acmart}

\usepackage{subcaption}
\usepackage{multirow}
\usepackage{colortbl}

\AtBeginDocument{%
  }

\acmYear{2026}
\setcopyright{cc}
\setcctype{by}
\acmConference[HUMA '26]{The 6th International Workshop on Human-centric Multimedia Analysis}{November 10--14, 2026}{Rio de Janeiro, Brazil}
\acmBooktitle{The 6th International Workshop on Human-centric Multimedia Analysis (HUMA '26), November 10--14, 2026, Rio de Janeiro, Brazil}
\acmDOI{10.1145/3841192.3841760}
\acmISBN{979-8-4007-2937-9/2026/11}

\begin{document}

\title{Tomatoes, Potatoes, and Onions: Questioning the Need for Faces in Face Presentation Attack Detection}
\author{Guray Ozgur}
\correspondingauthor
\orcid{0000-0002-1966-6641}
\affiliation{
  \institution{Fraunhofer IGD and Department of Computer Science, TU Darmstadt}
  \city{Darmstadt}
  \country{Germany}
}
\email{guray.ozgur@igd.fraunhofer.de}

\author{Fadi Boutros}
\orcid{0000-0003-4516-9128}
\affiliation{
  \institution{Fraunhofer IGD}
  \city{Darmstadt}
  \country{Germany}
}

\author{Naser Damer}
\orcid{0000-0001-7910-7895}
\affiliation{
  \institution{Fraunhofer IGD and Department of Computer Science, TU Darmstadt}
  \city{Darmstadt}
  \country{Germany}
}

\renewcommand{\shortauthors}{Guray Ozgur, Fadi Boutros and Naser Damer}

\begin{abstract}
Face presentation attack detection (PAD) is traditionally formulated as a face-specific problem, although many of the visual artifacts introduced by print, replay, and recapture processes are not inherently tied to facial appearance. In this work, we investigate whether transferable PAD representations can be learned without using faces during downstream PAD training. To this end, we introduce \textbf{TPO}, a controlled face-free presentation attack dataset consisting of bona fide, print, and replay recordings of, almost randomly chosen, tomatoes, potatoes, and onions acquired under protocols that closely mirror conventional face PAD datasets. Using a foundation-model-based PAD architecture, we demonstrate that a detector trained on TPO achieves an average AUC of \textbf{92.70\%} across four standard cross-dataset face PAD benchmarks, outperforming training on synthetic faces and remaining competitive with models trained on real face datasets. Conversely, models trained on face PAD datasets transfer consistently above chance to TPO, suggesting that the learned representations capture characteristics of the presentation process rather than object semantics. Furthermore, incorporating TPO into conventional face PAD training consistently improves cross-dataset performance under fixed optimization budgets, indicating that face-free data provides complementary information rather than simply additional training samples. Finally, representation and frequency analyses provide further evidence that transferable PAD representations cannot be explained by a single spectral artifact but instead encode richer presentation cues shared across object categories. Together, these results provide empirical evidence that transferable presentation attack representations can be learned independently of facial content, opening new opportunities for privacy-preserving and identity-independent PAD development.
\end{abstract}

\begin{teaserfigure}
  \centering
  \includegraphics[width=0.8\textwidth]{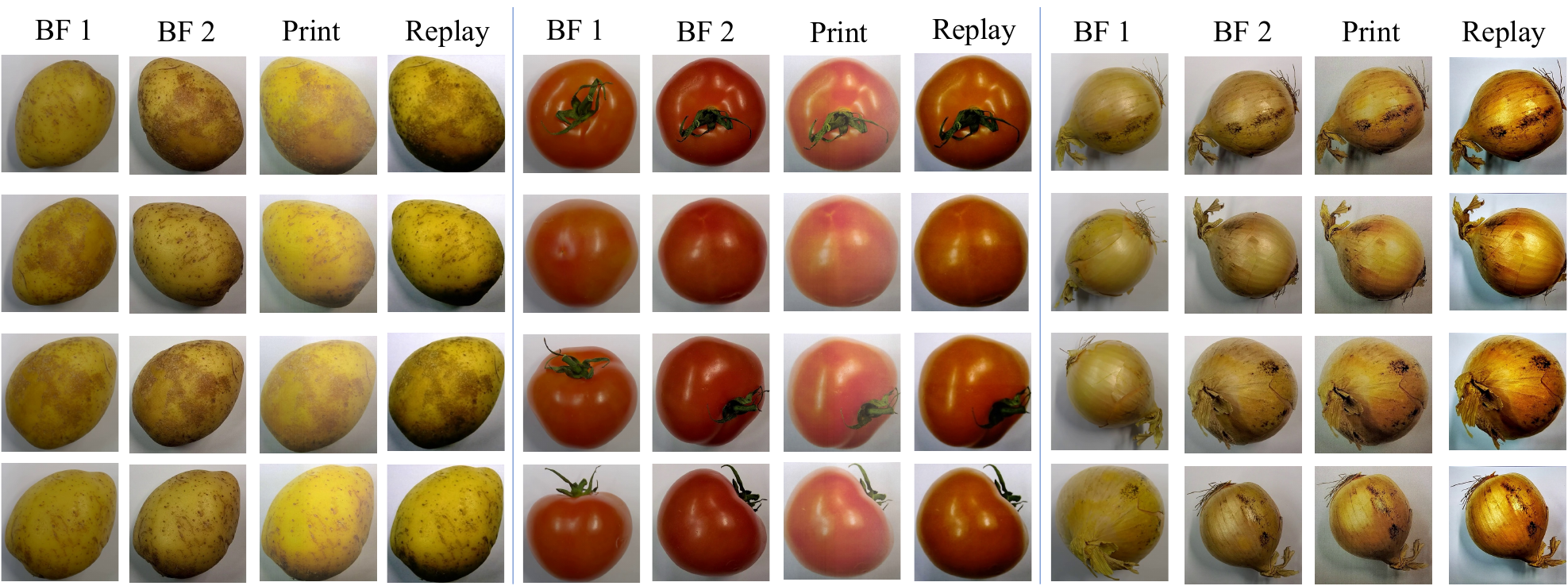}
  \includegraphics[width=0.8\textwidth]{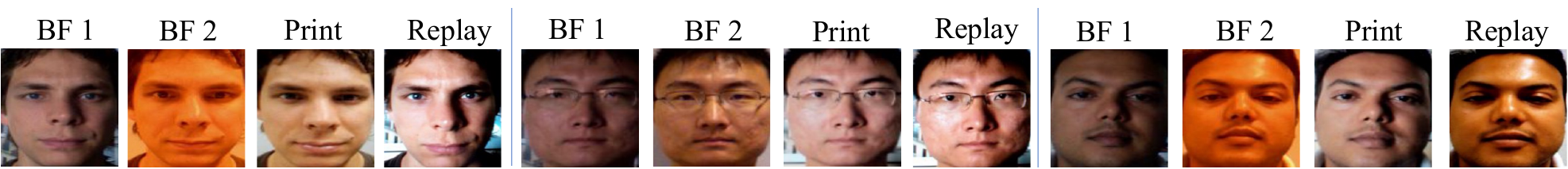}
  \caption{\textbf{Same presentation process, different objects.} TPO examples (top) and Idiap Replay-Attack examples (bottom) contain the same PAD structure: bona fide camera observations and print or display attacks recaptured by a second camera. Replacing faces with tomatoes, potatoes, and onions preserves the acquisition and presentation-instrument cues while removing facial identity, which makes the cross-object transfer test possible. The TPO dataset is released at \textcolor{magenta}{\url{https://github.com/gurayozgur/TPO}}.}
  \Description{Two horizontal sample grids compare the vegetable TPO dataset above with the face-based Idiap Replay-Attack dataset below. Each grid contains bona fide camera images and recaptured print or display attacks. The objects differ, but both grids show the same visual effects of printing, display replay, and camera recapture.}
  \label{fig:teaser}
\end{teaserfigure}


\maketitle
\balance
\section{Introduction}
\label{sec:intro}

Face recognition (FR) systems are deployed in increasingly high-stakes settings, from border control to financial authentication \cite{DBLP:conf/cvpr/DengGXZ19}, which makes them an attractive target for presentation attacks (PAs): a printout, a replayed video, or a mask presented to the camera in order to impersonate an enrolled identity or to evade recognition \cite{DBLP:conf/icb/ZhangYLLYL12, DBLP:conf/fgr/BoulkenafetKLFH17, DBLP:conf/eccv/ZhangYLYYSL20, DBLP:journals/pr/FangDKK22}. PAD, also called face anti-spoofing (FAS), is the countermeasure, and a decade of research has produced a rich literature of architectures, losses, and domain-generalization schemes aimed at separating bona fide presentations from attacks \cite{DBLP:journals/csur/RaghavendraB17,DBLP:journals/ijon/FathollahiPHK25}. Almost without exception, literature treats PAD as a \emph{face} problem. Models are trained on datasets of faces, benchmarked on cross-dataset face protocols such as MSU-MFSD \cite{DBLP:journals/tifs/WenHJ15}, CASIA-FASD \cite{DBLP:conf/icb/ZhangYLLYL12}, Idiap Replay-Attack \cite{DBLP:conf/biosig/ChingovskaAM12}, and OULU-NPU \cite{DBLP:conf/fgr/BoulkenafetKLFH17}. Their well-documented drop in cross-dataset performance is explained as a \emph{face-domain} shift: different subjects, demographics, illumination, and backgrounds. This framing has practical consequences. It motivates the collection of ever-larger face PAD corpora, raising consent \cite{DBLP:conf/cvpr/FangHD23}, privacy \cite{DBLP:conf/cvpr/FangHD23}, and demographic-bias \cite{DBLP:journals/pr/FangYKSD24} concerns because the data are images of real people, and it drives the design of face-specific architectures and auxiliary supervision such as facial depth or remote photoplethysmography.

In this position paper, we challenge the premise. We argue that a face PAD model does not, in any meaningful sense, necessarily need to learn faces. What it ideally learns is the signal introduced by the \emph{presentation attack instrument} (PAI): the moir\'e and sub-pixel structure of a display, the halftoning and paper texture of a print, the specular highlights, gamma distortion, and recapture noise that arise whenever an image is displayed or printed and then photographed again. These traces are properties of the \emph{recapture pipeline}, not of the object being recaptured. If this is correct, then the face is incidental: any object, presented bona fide and then attacked through the same instruments, should teach a detector the same thing. We test this claim in its strongest form. We assemble a PAD dataset that contains \emph{no faces at all}, bona fide and print/replay-attacked recordings of three classes of, for no specific reason, vegetables (tomatoes, potatoes, and onions), acquired under a face-PAD-style protocol, and we ask whether a detector trained only on these vegetables can detect attacks on human faces. It can. A foundation-model detector adapted purely on vegetables attains an average video-level Area under the Receiver Operating Characteristic curve (AUC) of $92.7\%$ across the four standard face benchmarks despite never having seen a face, compared on identical targets, it surpasses the same detector trained on a single real face dataset for two of the four sources and comes within one AUC point for a third. The reverse direction is weaker: common-budget single-source face models average $78.4\%$ AUC on the vegetables attacks, below their $89.3\%$ average on held-out face datasets but above the $67.7\%$ zero-shot result. Whether one trains on faces or on tomatoes is, for the purpose of the detector, tomayto-tomahto. Figure \ref{fig:teaser} makes the visual control explicit: the object changes, while the bona fide, print, replay, and recapture structure is retained. We then dissect the effect through controlled experiments. We analyze transferable ability.  We discuss the possibility of mixing face and non-face data. We look into model behavior under both face and non-PAD data, where representation and frequency analyses supports the statement: bona fide/attack organization is visible across object domains, while the six-domain frequency residual rules out a single global spectral shortcut.

Our main contributions are: (1) a \emph{position}, face PAD is more accurately understood as recapture-artifact detection, and the face is not what the detector necessarily learns. (2) \textbf{TPO}, a controlled, face-free vegetable PAD dataset ($12{,}480$ samples) that makes the claim testable. (3) a suite of controlled cross-object experiments, veg$\rightarrow$face, face$\rightarrow$veg, per-attack, per-object, data-scale, and a prior-knowledge axis, quantifying what transfers and what is required for it. To facilitate future research on privacy-preserving and identity-independent PAD, we publicly release the TPO dataset, together with the complete training, evaluation, and cross-dataset protocols.

\section{Background}
\label{sec:background}

\paragraph{Face PAD and cross-dataset generalization.}
Early face PAD relied on hand-crafted texture and frequency cues \cite{DBLP:conf/biosig/ChingovskaAM12,DBLP:conf/bmvc/DamerD16}, deep models later dominated intra-dataset benchmarks \cite{DBLP:conf/wacv/FangDKK22,DBLP:journals/pami/YuWQLLZ21,DBLP:journals/tbbis/YuLSXZ21,DBLP:conf/cvpr/LiuJ018}. The central open problem is cross-dataset generalization: a detector trained on one corpus degrades sharply on another, which the community attributes to domain shift and addresses through domain generalization and adaptation, adversarial alignment, meta-learning, and disentanglement \cite{DBLP:journals/tcsv/YanZH22}. Crucially, the ``domain'' in these works is implicitly the \emph{face} domain (subjects, sensors, environments), and evaluation is confined to face datasets. Our position re-frames part of this gap: maybe what shifts across datasets is the PAI and capture pipeline, not the face.

\paragraph{Reducing the data burden: synthetic and foundation models.}
Two recent directions aim to lessen the dependence on large, sensitive face PAD corpora. Synthetic face PAD data (e.g., SynthASpoof \cite{DBLP:conf/cvpr/FangHD23}) replaces real subjects with generated faces to sidestep privacy and consent, showing that \emph{identity} is not essential for PAD training. Even a competition on PAD development based on synthetic data stressed this possibility \cite{DBLP:conf/icb/FangHFRDAKPYHCZPJLSWLCZTSAS23}.  The other direction takes advantage of foundation models, CLIP and DINOv2 adapted with lightweight parameter-efficient tuning such as LoRA, have been shown to generalize strongly in low-data PAD regimes (e.g., FoundPAD \cite{DBLP:conf/wacv/OzgurCCBRD25}), and CLIP-based multimodal detectors (MMDA \cite{DBLP:conf/iccv/YangLYZLLYC25,DBLP:journals/corr/abs-2505-09484}) lead current cross-domain multimodal PAD. We build directly on these: SynthASpoof shows identity is unnecessary; we go one step further and show that the \emph{face itself} is unnecessary. We adopt a FoundPAD-style CLIP+LoRA detector as our foundation-model exemplar.

\paragraph{What does a PAD model actually use?}
A parallel line of work probes PAD decision cues, moir\'e patterns, face image quality \cite{DBLP:conf/cvpr/LiangQL22}, color/texture distortion \cite{DBLP:journals/tifs/WenHJ15}, reflections \cite{DBLP:journals/corr/abs-1902-10311}, and recapture noise \cite{DBLP:conf/eccv/JourablooLL18}, often via frequency analysis or saliency \cite{DBLP:conf/wacv/Cao025}. These are, by construction, properties of the recapture process. Our work turns this observation into a testable position: if the cue is the PAI trace, the presented object should be interchangeable. Vegetables, just as good as any other non-face object, provide a clean, face-free instrument for that test.

\section{The TPO Dataset}
\label{sec:dataset}
TPO is a face-free PAD dataset designed to preserve the acquisition factors of face PAD while replacing human subjects with non-human objects in order to question the need for actual face images to develop PAD solutions. It contains, for no specific reason, vegetables. Namely tomatoes, potatoes, and onions, with 26 physically distinct specimens per species and 78 object identities in total. These varieties of vegetables were selected as they have different levels of reflectiveness that might affect the developed PAD, at least theoretically. The same bona fide, print, and replay construction is applied to every species, making it possible to vary the presented object without changing the presentation process.

\paragraph{Bona fide acquisition.}
Each object was captured indoors under controlled illumination from four approximately orthogonal viewpoints, obtained by rotating the object between acquisitions. Every viewpoint was recorded at close and far scales with a Microsoft Surface tablet and a Samsung Galaxy smartphone. Reference markers on the display were used to keep object position stable while scale changed. For each combination of viewpoint, scale, and device, we captured one video of at least five seconds and one still image. Thus, each identity contributes 16 bona fide videos and 16 bona fide images, for 1,248 of each media type across the dataset.

\paragraph{Presentation attacks.}
Print attacks were produced from the bona fide still images, resized to fit an A4 sheet, printed on a standard office printer (Konica Minolta C450i with $600 \times 600$ dpi), and recaptured as videos. Replay attacks were produced by displaying the bona fide videos on either acquisition device and recording the display with either device. Replay capture was performed at a separate position in the same room with ambient illumination disabled to limit reflections. The replay recordings were segmented without re-encoding so that recapture compression remained intact. For both attack instruments, we exhaustively cross-source and capture device (Microsoft Surface or Samsung Galaxy), source and capture scale (close or far), object identity, and viewpoint. Each instrument therefore contributes 4,992 attack videos. This factorial protocol includes matched and mismatched device and scale conditions rather than binding an attack to a single capture pipeline. Samples of the dataset are shown in the upper part of Figure \ref{fig:teaser} and a statistical overview of the TPO data is presented in Table \ref{tab:dataset_statistics}.

\begin{table}[t]
\centering
\caption{\textbf{TPO composition.} TPO contains 12,480 presentations from 78 vegetable identities. Print and replay attacks are exactly balanced at 4,992 videos each; the bona fide subset contains 1,248 videos and 1,248 still images.}
\label{tab:dataset_statistics}
\begin{tabular}{lccc}
\hline
\textbf{Subset} & \textbf{Videos} & \textbf{Images} & \textbf{Total} \\
\hline
Bona fide & 1,248 & 1,248 & 2,496 \\
Print attacks & 4,992 & - & 4,992 \\
Replay attacks & 4,992 & - & 4,992 \\
\hline
\textbf{Total} & \textbf{11,232} & \textbf{1,248} & \textbf{12,480} \\
\hline
\end{tabular}
\end{table}

\section{Experimental Settings}
\label{sec:setup}

\paragraph{Datasets and cross-dataset protocol.}
We abbreviate MSU-MFSD \cite{DBLP:journals/tifs/WenHJ15}, CASIA-FASD \cite{DBLP:conf/icb/ZhangYLLYL12}, Idiap Replay-Attack \cite{DBLP:conf/biosig/ChingovskaAM12}, and OULU-NPU \cite{DBLP:conf/fgr/BoulkenafetKLFH17} as M, C, I, and O; their unweighted mean is MCIO. SynthASpoof \cite{DBLP:conf/cvpr/FangHD23} is the synthetic-face training control. We use the established FoundPAD \cite{DBLP:conf/wacv/OzgurCCBRD25} cropped-frame protocols for the face datasets. A dataset is used in its entirety as either a training source or a held-out target: no result trains and tests on the same dataset, and no train, development, or test partition is carved out of TPO. Single-source models are evaluated on the other three face datasets and TPO; multi-source models are evaluated on the omitted face dataset. This full-source cross-dataset design is the only protocol used for the results reported in the paper.

\paragraph{Frame construction and pre-processing.}
For each TPO video, we sample five frames uniformly over the middle 80\% of its duration, excluding the first and last 10\%s to avoid black margins that can remain after replay segmentation. Each still image contributes one frame. Frames are stored at $256{\times}256$, producing 57,408 training images from the 12,480 presentations. At model input, every TPO and face frame is converted to RGB and resized to $224{\times}224$ to match the model input size; no face detector or face-specific landmark operation is applied by our pipeline. Training augmentation follows the photometric protocol used by DADM/MMDA \cite{DBLP:conf/iccv/YangLYZLLYC25,DBLP:journals/corr/abs-2505-09484}: horizontal flip, additive Gaussian noise with standard deviation sampled from $[0,0.2]$, brightness shift up to $12\%$, independent RGB shifts up to 40 intensity levels, and gamma in $[0.5,1.5]$, each applied independently with probability 0.5. Evaluation uses only resizing and normalization.

\paragraph{FoundPAD and normalization correction.}
Our main detector follows FoundPAD \cite{DBLP:conf/wacv/OzgurCCBRD25}: This is motivated by the performance achieved by this foundation-model approach given limited domain data. The solution is a CLIP ViT-B/16 image encoder, rank-stabilized LoRA on the query and value projections of all 12 attention blocks ($r{=}8$, $\alpha{=}8$, dropout $0.4$), and a linear two-class head on the L2-normalized image embedding. The CLIP backbone is frozen; only the LoRA weights and head are optimized, totaling approximately 0.30M trainable parameters. The public FoundPAD data pipeline applies ImageNet normalization to a CLIP encoder. We instead use CLIP's native channel mean $(0.4815,0.4578,0.4082)$ and standard deviation $(0.2686,0.2613,0.2758)$. This correction removes a preprocessing mismatch with the statistics under which CLIP was pretrained, i.e., it does not alter the FoundPAD architecture. All FoundPAD results in the paper, including face-only baselines, are retrained with this correction.

The prior comparison uses a ViT-B/16 pretrained on ImageNet-21k and fine-tuned on ImageNet-1k. It receives ImageNet normalization and is fully fine-tuned, with the same resizing and augmentation pipeline, classification head, optimizer family, training data, and epoch budget as the TPO-trained FoundPAD model. This comparison controls the architecture family and downstream data, but not normalization or the parameter-efficient adaptation scheme. We therefore interpret it as evidence about the value of the CLIP representation rather than as a single-variable causal ablation.

\paragraph{Optimization and model selection.}
Following FoundPAD, we use inverse-class-frequency sampling and gradient clipping; following the DADM/MMDA training settings, we use the AdamW optimizer without a learning-rate schedule. Specifically, all trained models use two-class cross-entropy, AdamW with $\beta_1{=}0.9$, $\beta_2{=}0.999$, weight decay $5\times10^{-5}$, and gradient clipping at norm 5. The batch size is 48. For FoundPAD, the LoRA learning rate is $5\times10^{-6}$, and the head learning rate is $10^{-4}$, the ImageNet ViT uses the same backbone and head rates. In each epoch, the sampler requests as many draws as the source contains, with replacement and equal expected weight for bona fide and attack frames. The final incomplete batch is discarded. We use seed 777 throughout.

\paragraph{Training budgets and convergence.}
All experiments use a common training budget of approximately 11,000 optimizer updates (about 530,000 sampled image views with batch size 48) so that datasets of substantially different sizes are trained under the same computational budget. The budget was fixed a priori, before any paired comparisons or target-set evaluation, as a conservative amount of computation expected to allow convergence across all settings while remaining computationally practical. Rather than tuning the number of updates for individual datasets or methods, we intentionally keep the budget fixed to ensure that all observed performance differences arise from the training data rather than unequal optimization effort. For a source containing $N$ frames, we implement this budget as $\mathrm{clip}(\mathrm{round}(530000/N),2,60)$ complete epochs. Empirically, this budget is sufficient for convergence: every run finishes with training cross-entropy below 0.05, indicating that optimization has effectively converged within the allocated budget. Within the paired face experiments, the same exposure rule is applied to each face-only model and its face-plus-TPO counterpart. Under class-balanced sampling, bona fide and attack views retain equal expected weight, while samples within each class are drawn from the combined source pool. Consequently, TPO replaces a portion of the sampled real-face views rather than adding views on top of the original face exposure. This comparison therefore tests whether face-free recapture examples can improve a fixed face-training budget while reducing sampled real-face exposure.

\paragraph{Zero-shot and model analyses.}
The zero-shot baseline uses the frozen CLIP ViT-B/16 image and text encoders. We average L2-normalized text features from six paraphrases of each class, including the ISO/IEC 2382-37 \cite{ISO238237} terms ``bona-fide presentation'' and ``biometric presentation attack,'' and classify by the softmax of CLIP cosine similarities. We also analyze model behavior to understand in a deeper sense if, and to which degree, the TPO data do act similarly to face PAD data. For that, in Fig. \ref{fig:tsne}, both the TPO-trained and O-trained FoundPAD encoders receive the same 250 bona fide and 250 attack frames from each of M, C, I, O, TPO, and SynthASpoof. L2-normalized embeddings are reduced by PCA to 50 dimensions and then projected separately with t-SNE (perplexity 40, 1,500 iterations, fixed seeds). For Fig. \ref{fig:fft}, we sample 400 frames per class and dataset, convert them to grayscale, resize to $256{\times}256$, standardize each image, apply a two-dimensional Hann window, and radially average the log-magnitude FFT. Shading is the mean plus or minus $1.96$ standard errors. The attack residual is attack minus bona fide.

\paragraph{Metrics}
The network score is the softmax probability of the bona fide class. Frame scores belonging to the same source video are averaged before evaluation. A still image is treated as a one-frame presentation. We report video-level Area under the Receiver Operating Characteristic curve (AUC) and half total error rate (HTER), where $\mathrm{HTER}=(\mathrm{APCER}+\mathrm{BPCER})/2$ (where BPCER is the Bona-fide Presentation Classification Error Rate \cite{ISO301073} and APCER is the Attack Presentation Classification Error Rate  \cite{ISO301073}) at the target set's equal-error-rate threshold, following the common cross-dataset PAD convention. MCIO is the unweighted mean of the four target metrics.  

\begin{table*}[t]
\centering\scriptsize\setlength{\tabcolsep}{3.5pt}
\caption{\textbf{Cross-object transfer and the role of training source.} We compare ViT-B/16 models trained with no PAD data, TPO, SynthASpoof, or real face datasets, then evaluate on M, C, I, O, and TPO. TPO-trained FoundPAD reaches \textbf{92.70\% AUC} at \textbf{14.15\% HTER} on MCIO, compared with 67.81\% AUC for the architecture-matched ViT-B/16 initialized from ImageNet-21k and 81.02\% when FoundPAD is trained on SynthASpoof. Face-free transfer is therefore substantially stronger with the CLIP prior and exceeds the synthetic-face control. Single-source face-to-TPO transfer is uneven at 58.89-89.88\% AUC. HTER is measured at the test-EER threshold; gray cells denote training domains.}
\label{tab:main}
\resizebox{\textwidth}{!}{%
\begin{tabular}{ll|cc|cc|cc|cc||cc||cc}
\hline
\multicolumn{2}{c|}{\multirow{2}{*}{Training data / method}} & \multicolumn{2}{c|}{$\rightarrow$ M} & \multicolumn{2}{c|}{$\rightarrow$ C} & \multicolumn{2}{c|}{$\rightarrow$ I} & \multicolumn{2}{c||}{$\rightarrow$ O} & \multicolumn{2}{c||}{Avg.\ faces} & \multicolumn{2}{c}{$\rightarrow$ \textbf{TPO}} \\
\multicolumn{2}{c|}{} & HTER $\downarrow$ & AUC $\uparrow$ & HTER $\downarrow$ & AUC $\uparrow$ & HTER $\downarrow$ & AUC $\uparrow$ & HTER $\downarrow$ & AUC $\uparrow$ & HTER $\downarrow$ & AUC $\uparrow$ & HTER $\downarrow$ & AUC $\uparrow$ \\ \hline\hline
\multicolumn{14}{l}{\textit{No PAD training (zero-shot)}} \\
\multicolumn{2}{l|}{CLIP ViT-B/16, text-image prompts} & 46.67 & 53.52 & 52.78 & 44.44 & 37.45 & 66.58 & 42.12 & 59.89 & 44.75 & 56.11 & 36.89 & 67.72 \\
\hline
\multicolumn{14}{l}{\textit{Trained on \textbf{TPO} (vegetables, \emph{no faces})}} \\
\multicolumn{2}{l|}{\quad ViT-B/16, ImageNet-21k} & 41.90 & 59.98 & 48.56 & 55.76 & 21.35 & 85.29 & 36.07 & 70.22 & 36.97 & 67.81 & \cellcolor{black!10} & \cellcolor{black!10} \\
\multicolumn{2}{l|}{\quad ViT-B/16, CLIP+LoRA (FoundPAD)} & 15.00 & 91.01 & 11.11 & 95.98 & 12.20 & 94.08 & 18.30 & 89.74 & \textbf{14.15} & \textbf{92.70} & \cellcolor{black!10} & \cellcolor{black!10} \\
\hline
\multicolumn{14}{l}{\textit{Trained on SynthASpoof (synthetic faces)}} \\
\multicolumn{2}{l|}{\quad ViT-B/16, CLIP+LoRA (FoundPAD)} & 42.86 & 67.76 & 28.78 & 79.00 & 15.00 & 92.64 & 23.62 & 84.70 & 27.56 & 81.02 & 11.18 & 96.29 \\
\hline
\multicolumn{14}{l}{\textit{Trained on a single real face dataset}} \\
\multicolumn{2}{l|}{\quad ViT-B/16, CLIP+LoRA (FoundPAD) (train M)} & \cellcolor{black!10} & \cellcolor{black!10} & 5.22 & 98.97 & 9.50 & 97.09 & 9.73 & 95.05 & \cellcolor{black!10} & \cellcolor{black!10} & 19.58 & 87.85 \\
\multicolumn{2}{l|}{\quad ViT-B/16, CLIP+LoRA (FoundPAD) (train C)} & 24.05 & 83.79 & \cellcolor{black!10} & \cellcolor{black!10} & 21.85 & 84.27 & 24.66 & 84.24 & \cellcolor{black!10} & \cellcolor{black!10} & 46.53 & 58.89 \\
\multicolumn{2}{l|}{\quad ViT-B/16, CLIP+LoRA (FoundPAD) (train I)} & 28.57 & 81.82 & 35.67 & 73.07 & \cellcolor{black!10} & \cellcolor{black!10} & 16.76 & 89.75 & \cellcolor{black!10} & \cellcolor{black!10} & 30.32 & 76.97 \\
\multicolumn{2}{l|}{\quad ViT-B/16, CLIP+LoRA (FoundPAD) (train O)} & 20.24 & 90.14 & 4.67 & 98.86 & 9.55 & 94.91 & \cellcolor{black!10} & \cellcolor{black!10} & \cellcolor{black!10} & \cellcolor{black!10} & 19.31 & 89.88 \\
\hline
\multicolumn{14}{l}{\textit{Trained on several real face datasets}} \\
\multicolumn{2}{l|}{\quad ViT-B/16, CLIP+LoRA (FoundPAD) (train M\&I)} & \cellcolor{black!10} & \cellcolor{black!10} & 16.89 & 89.88 & \cellcolor{black!10} & \cellcolor{black!10} & 8.08 & 97.51 & \cellcolor{black!10} & \cellcolor{black!10} & 34.85 & 71.19 \\
\multicolumn{2}{l|}{\quad ViT-B/16, CLIP+LoRA (FoundPAD) (train O\&C\&I)} & 10.95 & 92.94 & \cellcolor{black!10} & \cellcolor{black!10} & \cellcolor{black!10} & \cellcolor{black!10} & \cellcolor{black!10} & \cellcolor{black!10} & \cellcolor{black!10} & \cellcolor{black!10} & 25.60 & 82.92 \\
\multicolumn{2}{l|}{\quad ViT-B/16, CLIP+LoRA (FoundPAD) (train O\&M\&I)} & \cellcolor{black!10} & \cellcolor{black!10} & 15.00 & 94.95 & \cellcolor{black!10} & \cellcolor{black!10} & \cellcolor{black!10} & \cellcolor{black!10} & \cellcolor{black!10} & \cellcolor{black!10} & 32.77 & 73.99 \\
\multicolumn{2}{l|}{\quad ViT-B/16, CLIP+LoRA (FoundPAD) (train O\&C\&M)} & \cellcolor{black!10} & \cellcolor{black!10} & \cellcolor{black!10} & \cellcolor{black!10} & 18.30 & 89.38 & \cellcolor{black!10} & \cellcolor{black!10} & \cellcolor{black!10} & \cellcolor{black!10} & 30.35 & 75.73 \\
\multicolumn{2}{l|}{\quad ViT-B/16, CLIP+LoRA (FoundPAD) (train I\&C\&M)} & \cellcolor{black!10} & \cellcolor{black!10} & \cellcolor{black!10} & \cellcolor{black!10} & \cellcolor{black!10} & \cellcolor{black!10} & 19.12 & 88.00 & \cellcolor{black!10} & \cellcolor{black!10} & 32.63 & 72.86 \\
\hline
\end{tabular}}
\end{table*}

\begin{table*}[t]
\centering\scriptsize\setlength{\tabcolsep}{3pt}
\caption{\textbf{Can TPO replace part of single-source face exposure?} Replacing part of a fixed single-source face-training budget with TPO consistently improves cross-dataset PAD. Under identical optimization budgets, introducing face-free presentation data increases mean AUC from 89.33\% to 92.55\% while reducing mean HTER from 17.54\% to 13.97\%, showing that TPO contributes complementary presentation information rather than simply increasing training data..}
\label{tab:e9single}
\resizebox{0.8\textwidth}{!}{%
\begin{tabular}{ll|cccccccccccc||c}
\hline
Training data & Metric & C$\to$I & C$\to$M & C$\to$O & I$\to$C & I$\to$M & I$\to$O & M$\to$C & M$\to$I & M$\to$O & O$\to$I & O$\to$M & O$\to$C & Avg. \\ \hline\hline
\multirow{2}{*}{face only} & HTER $\downarrow$ & 21.85 & 24.05 & 24.66 & 35.67 & 28.57 & 16.76 & 5.22 & 9.50 & 9.73 & 9.55 & 20.24 & 4.67 & 17.54 \\
 & AUC $\uparrow$ & 84.27 & 83.79 & 84.24 & 73.07 & 81.82 & 89.75 & 98.97 & 97.09 & 95.05 & 94.91 & 90.14 & 98.86 & 89.33 \\ \hline
\multirow{2}{*}{\textbf{face + TPO}} & HTER $\downarrow$ & 15.00 & 16.90 & 21.62 & 25.33 & 25.00 & 18.59 & 0.67 & 5.55 & 5.95 & 13.10 & 17.14 & 2.78 & \textbf{13.97} \\
 & AUC $\uparrow$ & 91.93 & 89.02 & 87.28 & 86.55 & 84.73 & 88.65 & 99.90 & 98.89 & 97.95 & 94.15 & 92.07 & 99.43 & \textbf{92.55} \\ \hline
\end{tabular}}
\end{table*}

\begin{table*}[t]
\centering\scriptsize\setlength{\tabcolsep}{3pt}
\caption{Face-free data remains beneficial even in stronger multi-source training regimes. Replacing part of the sampled face observations with TPO improves average AUC from 92.11\% to 96.96\% and reduces HTER from 14.72\% to 7.84\%, indicating that the benefit of face-free presentation data extends beyond low-data scenarios.}
\label{tab:e9multi}
\resizebox{0.7\textwidth}{!}{%
\begin{tabular}{ll|cc|cccc||c}
\hline
\multirow{2}{*}{Training data} & \multirow{2}{*}{Metric} & \multicolumn{2}{c|}{double-source} & \multicolumn{4}{c||}{triple-source} & \multirow{2}{*}{Avg.} \\
 & & M\&I$\to$C & M\&I$\to$O & O\&C\&I$\to$M & O\&M\&I$\to$C & O\&C\&M$\to$I & I\&C\&M$\to$O & \\ \hline\hline
\multirow{2}{*}{face only} & HTER $\downarrow$ & 16.89 & 8.08 & 10.95 & 15.00 & 18.30 & 19.12 & 14.72 \\
 & AUC $\uparrow$ & 89.88 & 97.51 & 92.94 & 94.95 & 89.38 & 88.00 & 92.11 \\ \hline
\multirow{2}{*}{\textbf{face + TPO}} & HTER $\downarrow$ & 3.44 & 8.06 & 16.19 & 2.44 & 11.00 & 5.88 & \textbf{7.84} \\
 & AUC $\uparrow$ & 99.11 & 96.95 & 92.07 & 99.50 & 95.82 & 98.29 & \textbf{96.96} \\ \hline
\end{tabular}}
\end{table*}

\begin{table*}[t]
\centering\scriptsize\setlength{\tabcolsep}{3.5pt}
\caption{\textbf{What makes face-free training transfer?} Attack diversity is more important than dataset size for face-free PAD training. Restricting TPO to a single attack type substantially reduces transfer performance, whereas individual vegetable categories already provide meaningful generalization. Combining multiple vegetable types yields the strongest results, while using only a small fraction of available video frames causes almost no performance degradation, suggesting that presentation diversity rather than redundant frames drives transfer.}
\label{tab:e3}
\resizebox{0.9\textwidth}{!}{%
\begin{tabular}{ll|cc|cc|cc|cc||cc}
\hline
\multicolumn{2}{c|}{\multirow{2}{*}{TPO training subset}} & \multicolumn{2}{c|}{$\rightarrow$ M} & \multicolumn{2}{c|}{$\rightarrow$ C} & \multicolumn{2}{c|}{$\rightarrow$ I} & \multicolumn{2}{c||}{$\rightarrow$ O} & \multicolumn{2}{c}{Average} \\
\multicolumn{2}{c|}{} & HTER $\downarrow$ & AUC $\uparrow$ & HTER $\downarrow$ & AUC $\uparrow$ & HTER $\downarrow$ & AUC $\uparrow$ & HTER $\downarrow$ & AUC $\uparrow$ & HTER $\downarrow$ & AUC $\uparrow$ \\ \hline\hline
\multicolumn{2}{l|}{all vegetables, both PAIs (reference)} & 15.00 & 91.01 & 11.11 & 95.98 & 12.20 & 94.08 & 18.30 & 89.74 & \textbf{14.15} & \textbf{92.70} \\
\multicolumn{2}{l|}{print attacks only} & 27.38 & 81.26 & 8.67 & 97.26 & 23.10 & 86.04 & 28.86 & 80.00 & 22.00 & 86.14 \\
\multicolumn{2}{l|}{replay attacks only} & 28.81 & 80.88 & 26.00 & 83.47 & 17.50 & 89.72 & 26.48 & 81.81 & 24.70 & 83.97 \\
\multicolumn{2}{l|}{tomatoes only} & 24.05 & 82.54 & 14.11 & 92.60 & 27.60 & 80.62 & 29.28 & 76.79 & 23.76 & 83.14 \\
\multicolumn{2}{l|}{potatoes only} & 25.95 & 80.94 & 27.33 & 82.02 & 25.55 & 84.26 & 24.13 & 84.39 & 25.74 & 82.90 \\
\multicolumn{2}{l|}{onions only} & 19.76 & 88.78 & 12.44 & 95.04 & 12.30 & 94.77 & 19.71 & 88.54 & 16.05 & 91.78 \\
\multicolumn{2}{l|}{50\% of frames} & 15.71 & 90.57 & 11.33 & 95.70 & 12.55 & 94.02 & 20.38 & 87.58 & 14.99 & 91.97 \\
\multicolumn{2}{l|}{25\% of frames} & 15.95 & 91.30 & 9.22 & 97.23 & 14.15 & 92.77 & 18.09 & 89.53 & 14.35 & 92.71 \\
\multicolumn{2}{l|}{10\% of frames} & 13.57 & 91.47 & 9.56 & 96.88 & 13.50 & 93.48 & 17.60 & 90.03 & \textbf{13.56} & \textbf{92.97} \\
\hline
\end{tabular}}
\end{table*}

\section{Position and Empirical Evidence}
\label{sec:evidence}

The central hypothesis investigated in this work is that face presentation attack detection does not necessarily depend on faces for its development. Rather than learning semantic facial characteristics, a PAD detector may primarily learn visual artifacts introduced by the presentation attack instrument (PAI) and the recapture process. If this hypothesis holds, a detector trained using only non-face objects should generalize to face PAD, while a detector trained on face PAD should also exhibit transfer to non-face presentation attacks. Furthermore, if presentation artifacts rather than facial appearance are the dominant learning signal, face-free data should complement conventional face PAD datasets during training. The following experiments evaluate this hypothesis through cross-object transfer, comparisons with synthetic and real face datasets, analysis of representation priors, controlled training-data replacement experiments, and ablation studies.

\subsection{Face-free training transfers to face PAD}

The primary evidence supporting our position is presented in Table \ref{tab:main}. Training FoundPAD exclusively on the proposed TPO dataset, without using any human face during downstream training, achieves 92.70\% average AUC and 14.15\% HTER across the four standard cross-dataset face PAD benchmarks (MCIO). These results demonstrate that learning from a completely face-free dataset is not only feasible, but highly competitive. When compared with conventional single-source face PAD training, TPO training achieves comparable or superior performance on most evaluation protocols despite containing no facial information. This directly supports our hypothesis that the presentation process itself provides sufficient supervision for learning transferable PAD representations. Interestingly, even when compared with substantially stronger training regimes based on multiple combined face datasets, TPO remains competitive. Although pooled face datasets achieve the highest overall performance, the relatively small performance gap indicates that much of the transferable PAD knowledge can already be learned from the non-face presentation process alone. Table \ref{tab:main} also includes a comparison with SynthASpoof. Despite containing synthetic faces specifically designed for PAD training, SynthASpoof reaches only 81.02\% average AUC, considerably below the 92.70\% obtained using TPO. This suggests that preserving realistic presentation attack artifacts is more important than preserving facial appearance. The reverse transfer direction provides additional evidence. Models trained on individual face PAD datasets achieve between 58.89\% and 89.88\% AUC when evaluated on TPO, consistently outperforming the CLIP zero-shot baseline in nearly every case. Although transfer from faces to vegetables is weaker than the reverse direction, these results remain substantially above random performance and indicate that the learned representations capture characteristics of the attack process rather than object semantics. Finally, Table \ref{tab:main} compares FoundPAD against an architecture-matched ViT-B/16 initialized from ImageNet-21k. Under identical downstream training data, the ImageNet-pretrained model reaches only 67.81\% average AUC, whereas FoundPAD achieves 92.70\%. This large performance difference highlights the importance of the representation prior provided by the foundation model. Consequently, successful face-free transfer depends not only on the training data but also on sufficiently general visual representations capable of capturing presentation artifacts.

\subsection{Face-free data complements face PAD training}

The previous experiment demonstrated that face-free training alone produces competitive PAD performance. We now investigate whether TPO also provides complementary information when combined with conventional face datasets. Table \ref{tab:e9single} evaluates replacing part of the fixed face-training budget with TPO while keeping the total optimization budget unchanged. Across the twelve single-source cross-dataset protocols, introducing TPO improves performance in the majority of cases, increasing the average AUC from 89.33\% to 92.55\% while reducing the average HTER from 17.54\% to 13.97\%. These improvements are particularly noteworthy because TPO does not increase the overall amount of optimization or total number of sampled training examples. Instead, a portion of real-face observations is replaced by non-face presentation data. The consistent improvements therefore indicate that TPO contributes complementary information regarding presentation artifacts that is not fully represented by face-only training data. Table \ref{tab:e9multi} extends this analysis to stronger multi-source training regimes. Even when two or three face datasets are already available, replacing part of the sampled face observations with TPO continues to improve performance. Average AUC increases from 92.11\% to 96.96\%, while HTER decreases from 14.72\% to 7.84\%. These results demonstrate that the benefit of TPO is not limited to low-data settings, but remains observable even when substantial face PAD supervision is already available. Overall, these findings indicate that presentation artifacts learned from face-free data complement rather than duplicate the information contained in existing face PAD datasets.

\subsection{What properties of TPO matter?}

Table \ref{tab:e3} investigates which characteristics of TPO are responsible for the observed transfer.

Restricting training to only print attacks or only replay attacks substantially reduces performance, lowering the average AUC from 92.70\% to 86.14\% and 83.97\%, respectively. This suggests that diversity of presentation attack instruments is more valuable than increasing the number of samples from a single attack type. Training on individual vegetable categories also remains surprisingly effective. Each vegetable independently produces meaningful transfer far above random performance, with onions achieving 91.78\% average AUC. Nevertheless, combining tomatoes, potatoes, and onions consistently produces the strongest overall results. A plausible explanation is that the three object categories introduce complementary appearance characteristics, including different surface reflectance, texture, and shape, resulting in richer presentation variability while preserving identical attack mechanisms. Interestingly, reducing the number of sampled frames has almost no effect on performance. Even when only 10\% of all available frames are used, performance remains essentially unchanged (92.97\% AUC, 13.56\% HTER). Since the vegetables remain largely static during acquisition, consecutive video frames contain highly redundant visual information. Consequently, increasing the number of nearly identical frames contributes little additional supervision, whereas diversity in presentation conditions appears substantially more important.

\begin{figure*}[t]
\centering
\includegraphics[width=0.7\textwidth]{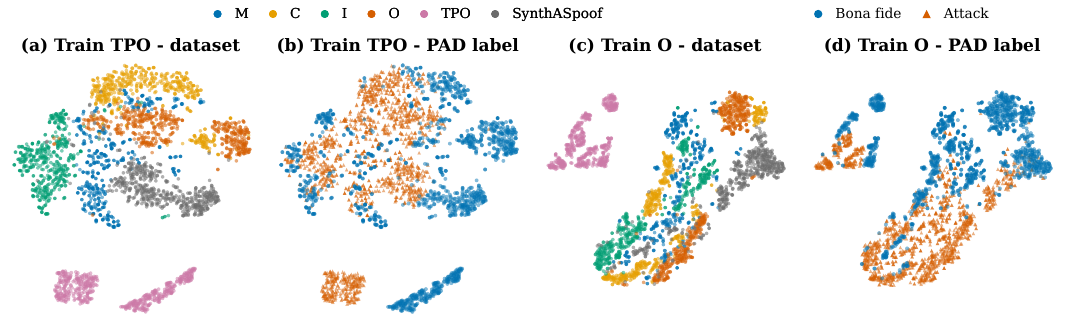}
\caption{\textbf{Training source changes the embedding geometry.} FoundPAD visual embeddings for the same class-balanced samples from M, C, I, O, TPO, and SynthASpoof. Panels (a-b) use the TPO-trained encoder; panels (c-d) use the O-trained encoder. Dataset coloring in (a,c) shows stronger source-domain organization after TPO training, while PAD coloring in (b,d) shows a clearer cross-domain bona fide/attack partition after O training. Each dataset contributes 250 bona fide and 250 attack images. Each encoder is projected separately with fixed-seed PCA-t-SNE; axes are arbitrary.}
\Description{Four scatter plots show identical images embedded by two FoundPAD encoders. In panels a and c, six colors identify M, C, I, O, TPO, and SynthASpoof. In panels b and d, blue circles mark bona fide images and orange triangles mark attacks. The TPO-trained pair forms stronger dataset-specific groups, whereas the O-trained pair shows a clearer separation between bona fide and attack points across datasets.}
\label{fig:tsne}
\end{figure*}

\subsection{Representation analysis}

\begin{figure*}[t]
\centering
\includegraphics[width=0.7\textwidth]{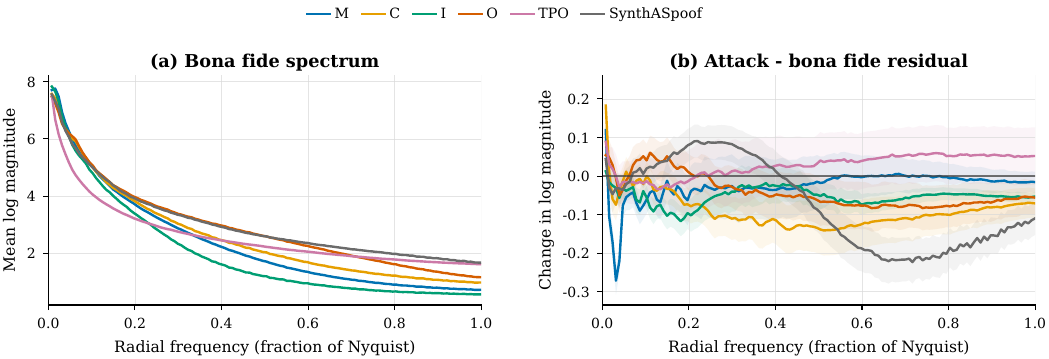}
\caption{\textbf{The transferable cue is not one global frequency signature.} Radial log-magnitude spectra for M, C, I, O, TPO, and SynthASpoof. Bona fide curves in (a) differ strongly by dataset and crop. Attack - bona fide residuals in (b) also disagree in direction: TPO gains high-frequency energy, C, I, O, and SynthASpoof lose it, and M remains near zero. Curves average 400 images per class and dataset; shading gives 95\% confidence intervals. Colors match Fig. \ref{fig:tsne}. Images are grayscale, resized to $256{\times}256$, standardized per image, and Hann-windowed before the FFT.}
\Description{Two line plots share six dataset colors. The left plot shows widely separated bona fide frequency spectra. The right plot shows attack minus bona fide residuals around a horizontal zero line: the TPO curve is positive at high frequencies, the C, I, O, and SynthASpoof curves are negative, and the M curve stays near zero. Narrow shaded bands show confidence intervals.}
\label{fig:fft}
\end{figure*}

The quantitative results are further supported by representation and frequency analyses shown in Figures \ref{fig:tsne} and \ref{fig:fft}. Figure \ref{fig:tsne} visualizes the embedding space learned after adaptation on TPO and, for comparison, after adaptation on OULU-NPU. When the encoder is trained on TPO, samples remain organized according to their originating datasets, indicating that object-domain information is preserved. Nevertheless, bona fide and attack presentations exhibit a consistent separation across all datasets, demonstrating that presentation status forms a transferable representation that generalizes beyond object identity. Training on a face dataset produces an even clearer separation between bona fide and attack samples, while preserving the same overall organization. Together, these observations indicate that PAD representations simultaneously encode dataset-specific appearance and presentation-specific characteristics, with the latter transferring across fundamentally different object categories. Figure \ref{fig:fft} investigates whether this transfer can be explained by a universal frequency signature. Although bona fide spectra differ substantially across datasets, the attack-minus-bona-fide residual exhibits inconsistent behavior: TPO attacks introduce additional high-frequency energy, whereas several face datasets show the opposite trend, and others remain approximately neutral. Therefore, the transferable PAD representation cannot be reduced to a single universal spectral artifact. Instead, the results suggest that the learned representation captures a richer combination of presentation cues than simple global frequency statistics. Together with the quantitative experiments, these analyses support the proposed view that face PAD primarily learns transferable characteristics of the presentation process rather than object-specific facial information.

\section{Counterarguments}
\label{sec:counterarguments}

We address the strongest objections to our position.

\paragraph{``This is just PAI detection, of course it is object-independent; it is not surprising.''}
We agree that, stated abstractly, the recapture-artifact view is intuitive, and prior work has gestured at moir\'e and recapture cues. The surprise is not conceptual but consequential: the entire face PAD ecosystem, datasets, benchmarks, privacy debates, demographic-bias studies, and face-specific architectures, is predicated on the face being central. If the face is inessential, that premise, and the practices built on it, deserve revision. Our contribution is to convert an intuition into a falsifiable, quantified claim (a detector that has literally never seen a face in its adaptation reaching $92.7\%$ AUC on faces) and to trace its boundary conditions.

\paragraph{``CLIP has seen faces, prints, and screens during pretraining, so the model is not really face-free.''}
Two responses. First, our claim concerns the \emph{downstream training data}: no faces are used to teach the PAD task, which is where privacy and bias costs are incurred. Second, our architecture-matched comparison shows that an ImageNet-pretrained ViT, which has also encountered faces and screens, transfers far less effectively than CLIP. Mere pretraining exposure to those contents is therefore insufficient to explain the result. What matters is representation generality, not face exposure.

\paragraph{``Vegetables are a toy; real systems need faces in the pipeline anyway.''}
PAD is a modular pre-filter to face recognition; our claim is about what the PAD module must be \emph{trained} on, not about removing faces from the overall system. Nothing in our position prevents deploying the detector in front of a face recognizer. Put simply: the \emph{training data} for the PAD module need not contain faces.

\paragraph{``Calibration does not transfer, so the result is weaker than it looks.''}
Threshold miscalibration across domains is universal in cross-dataset PAD, including face$\rightarrow$face, and is handled by target-domain threshold estimation. It does not affect the ranking/discrimination that AUC measures, which is the quantity relevant to our position.

\section{Implications}
\label{sec:implications}

If PAD is recapture-artifact detection, several consequences follow:

\textbf{Privacy-preserving and consent-free data.}
The most sensitive ingredient in PAD research, images of real people, with the attendant consent, GDPR, and retention burdens, is, for training purposes, replaceable. PAD training corpora can be built from non-human, non-identifiable objects recaptured through the target instruments, eliminating a class of privacy risk at its source. Vegetables are one instance; any textured object would do.

\textbf{Demographic fairness by construction.}
Demographic bias in PAD has been linked to imbalanced human training data \cite{DBLP:journals/pr/FangYKSD24}. A face-free training set has no demographic attributes to be imbalanced over, offering a route to detectors whose \emph{learned} decision cue is provably independent of skin tone, age, or gender. Fairness then reduces to ensuring PAI and sensor coverage rather than subject coverage. However, this stays theoretical as analyzing this was not part of this work.

\textbf{Rethinking the ``generalization gap''.}
Cross-dataset degradation in face PAD is routinely attributed to face-domain shift. Our results suggest a large part of it is PAI and capture-pipeline shift. Benchmarks and domain-generalization methods should disentangle these: a detector that generalizes across faces but not across printers has not solved the problem the field thinks it has. Reporting attack-instrument and sensor provenance may matter more than reporting subject demographics.

\textbf{Method design and evaluation.}
The prior-knowledge axis indicates that investment is best directed at general visual representations and at breadth of recapture conditions, rather than at face-specific inductive biases. Face-free sets like TPO provide a clean, cheaply extensible testbed for stress-testing PAD generalization and for auditing what a detector actually uses.

\textbf{A constructive research program.}
The position invites building intentionally diverse, face-free ``artifact'' corpora spanning many objects, printers, displays, and cameras; characterizing which physical recapture properties (moir\'e, halftoning, specularity, noise) drive transfer; and testing whether the account extends to modalities beyond RGB (depth, IR) and to attacks beyond print/replay (3D masks), where object-specific 3D structure may finally make the object matter.

\section{Conclusion}
\label{sec:conclusion}
This paper investigated a fundamental question in face presentation attack detection: \emph{does learning transferable PAD representations necessarily require faces?} To answer this question, we introduced TPO, a controlled face-free presentation attack dataset that preserves the acquisition and recapture process of conventional face PAD while removing facial content entirely. Our experiments consistently support the hypothesis that transferable PAD knowledge is primarily associated with presentation artifacts rather than facial appearance. A detector adapted exclusively on TPO achieved competitive cross-dataset face PAD performance, outperforming training on synthetic faces and approaching models trained on substantially larger collections of real face datasets. Conversely, models trained on face PAD datasets generalized consistently to TPO, demonstrating that the learned representations transfer across fundamentally different object categories. Moreover, replacing part of the real-face training data with TPO consistently improved performance in both single-source and multi-source training settings under identical optimization budgets, indicating that face-free presentation data contributes complementary information beyond conventional face datasets. Our ablation studies further suggest that diversity of presentation attack conditions is more important than increasing the number of nearly redundant observations, while the comparison with an ImageNet-pretrained backbone highlights the importance of strong general visual representations for enabling cross-object transfer. Finally, embedding and frequency analyses indicate that PAD representations organize samples according to presentation status across domains, yet cannot be explained by a single universal frequency signature, suggesting that the learned representations capture a richer set of presentation cues. We do not argue that faces are irrelevant for every aspect of biometric systems, nor that all presentation attack scenarios are object-independent. Instead, our findings provide empirical evidence that a substantial portion of transferable PAD knowledge can be learned without facial data. We hope this work motivates a broader rethinking of how PAD datasets are collected and how generalization is evaluated, encouraging future research toward privacy-preserving, identity-independent, and presentation-centered approaches to biometric presentation attack detection. On the limitation side, the conclusions drawn in this work are intentionally restricted to print and replay presentation attacks in the visible spectrum. Whether similar observations hold for attack types that depend more strongly on object geometry, such as three-dimensional masks, or for other sensing modalities including depth and infrared imaging, remains an open question. Furthermore, although TPO removes facial content from downstream PAD training, the employed CLIP foundation model has been pretrained on large-scale Internet imagery, including human faces. Our experiments demonstrate that such pretraining alone is insufficient to explain the observed transfer, yet understanding the minimal representation prior required for successful face-free PAD remains an important direction for future work.

\begin{acks}
This research work has been funded by the German Federal Ministry of Education and Research and the Hessian Ministry of Higher Education, Research, Science and the Arts within their joint support of the National Research Center for Applied Cybersecurity ATHENE.
\end{acks}

\section*{Ethics and Privacy Statement}
TPO contains no human subjects, biometric data, or personally identifiable information. The face datasets used for comparison were obtained from their original owners under the appropriate licenses. Our finding that PAD representations transfer without faces can reduce privacy risks by enabling face PAD development with substantially less collection and use of sensitive biometric data. Face PAD contributes to the security of biometric and convenience applications against presentation attacks. Our experiments focus on investigating this transferability and do not necessarily aim to demonstrate operational readiness. Since CLIP was pretrained on web imagery that may contain faces, we claim reduced collection of sensitive data rather than a fully face-free system. As the released data implements print and replay attacks already standard in public PAD research, we consider the additional misuse potential minimal.

\clearpage
\bibliographystyle{ACM-Reference-Format}
\bibliography{sample-base}

\end{document}